# Convolutional Spiking Neural Network for Image Classification

Mikhail Kiselev
*Dept. of future technologies*
*Kaspersky*
Moscow, Russia
mikhail.kiselev@kaspersky.com

Andrey Lavrentyev
*Dept. of future technologies*
*Kaspersky*
Moscow, Russia
andrey.lavrentyev@kaspersky.com

*Abstract*—We consider an implementation of convolutional architecture in a spiking neural network (SNN) used to classify images. As in the traditional neural network, the convolutional layers form informational "features" used as predictors in the SNN-based classifier with CoLaNET architecture. Since weight sharing contradicts the synaptic plasticity locality principle, the convolutional weights are fixed in our approach. We describe a methodology for their determination from a representative set of images from the same domain as the classified ones. We illustrate and test our approach on a classification task from the NEOVISION2 benchmark.

## I. INTRODUCTION

The tremendous progress of artificial intelligence technologies based on neural networks is evident to everybody. But, despite the new achievements of large neural networks appearing every month, the problems limiting further technological growth in this direction become clear. The most severe of them is unacceptable level of energy consumption necessary to train and use future even larger neural network models [1]. On the contrary, the biological brain demonstrates striking economy in this sense – tens of Watts instead of tens of megaWatts consumed by supercomputers used to train modern large linguistic models. Therefore, spiking neural network (SNN) which more closely mimic biological neural networks are quickly gaining in popularity. The event-based ideology of SNN emulation, very simple and asynchronous behavior of spiking neurons have been shown to reduce energy consumption by orders of magnitude when implemented on specialized neuromorphic hardware. However, SNNs often show poorer results in terms of accuracy in comparison with traditional artificial neural networks (ANNs) due to very efficient architectures used in the latter. For example, high accuracy in image classification tasks is achieved by ANNs due to use of convolutional architecture. This has inspired the research aimed to implementation of convolutional structures in SNN [2 - 4]. Examples of application of convolutional SNNs in various areas such as computer vision [5], speech recognition [6], hand-gesture recognition [7, 8] and even for text classification [9] have been reported.

Marrying the concepts of SNN and convolutional networks has two aspects – related to learning and inference. The main problem is rooted in the former, because inference in convolutional spiking networks follows the same principles as in ANNs – the convolution matrix is represented as a matrix of synaptic weights of the spiking neurons belonging to convolutional layer. This weight matrix is the same for all neurons on one map layer and convolution mechanism works in its usual way – neuron reacts to specific structural feature determined by the weight matrix (the convolution kernel). Learning the weight matrix becomes a problem in case of synaptic plasticity locality requirement. It imposes the condition on synaptic plasticity rules stating that synaptic weight changes can depend only on properties and activities of pre- and post-synaptic neurons. It is an important requirement because it is biologically plausible and adheres the principle of totally asynchronous SNN operation. However, the learning of convolutional weights is based on weight sharing – the corresponding weights of all neurons from one map layer should change equally that apparently contradicts the locality principle. In the majority of approaches [2 - 4, 6, 10] the locality principle is ignored and the convolution matrix is created as a result of unsupervised learning process. Some works [5, 11] use for obtaining the convolutional matrix the classic error backpropagation algorithm leveraged for SNN – the so called surrogate gradient method. This idea is further developed and applied to recurrent network in the SLAYER algorithm [12] used for training convolutional SNN in [7]. At last, convolutional SNN may be a result of the ANN-to-SNN conversion procedure where non-plastic SNN equivalent to pre-trained convolutional ANN is built and used for inference [13].

In contrast with the aforementioned works, our study is based on the following two assumptions:

- **No weight sharing.** All our research projects are oriented to implementation on neuromorphic processors like Loihi [14] or AltAI [15]. Weight sharing can hardly be implemented efficiently on this hardware platform since it would require unacceptably high traffic.

- **Domain-level convolutions.** Convolutional matrices should correspond to structural features characteristic for various images from the domain where the trained network is planned to use. Therefore, they can be common for different particular classification tasks in this domain. Of course, this assumption can hardly be proven formally but it follows from common sense.

The similar assumption is used in the L2L SNN learning methodology [16].

It follows from these assumptions that values of the convolution matrices can and should be obtained using some computational procedure on a sufficiently representative sample of images from the target domain and after that converted to non-plastic network structures in the SNNs solving various classification tasks from this domain. The learning itself is implemented in the higher plastic layer(s) of these SNNs. In the present paper we consider this procedure (Section 2). It is based on the ideas from [10] but further developed and represented as a standalone algorithm (in [10] it is implemented inside SNN). In Section 3 we consider formation of convolutional SNN layer from the convolutional matrices obtained. The architecture of the whole learning SNN and its application to object classification task from the Neovision2 benchmark is described in Section 4. In Section 5, we discuss the results obtained. After it, Conclusions follow.

## II. OBTAINING THE DOMAIN-SPECIFIC CONVOLUTIONS

Convolution kernels are usually square matrices of size $K \times K$ pixels which should correspond to typical structural elements of the images to be analyzed by neural networks when they are used for image classification, object detection etc. As it was said in Introduction, we determine the convolutions not for particular task but for all tasks from the given domain. To obtain them, we need a big collection of various images which represent many various samples from the domain of interest. This set **G** is input data for our convolution construction algorithm. It is assumed that they are monochrome images – extension to RGB images is quite straightforward.

Besides that, the algorithm hyperparameters should be specified. They are few; later, we discuss how to select them. They are:

$K$ – the convolution size;
$N_C$ – the convolution kernel count;
$s$ – the convolution stride;
$B$ – the pixel brightness threshold;
$w_{min}$ – the lower limit for convolution matrix elements (it is non-positive);
$w_{max}$ – the higher limit of convolution matrix elements;
$l$ – the learning rate.

The algorithm is iterative. Every iteration corresponds to one image in **G**. The result of this algorithm is the set of convolution kernels $\mathbf{w} = w_{aij}$, where $0 \leq a < N_C$, $0 \leq i, j < K$. The algorithm uses the auxiliary tensor $\mathbf{W} = W_{aij}$ of the same shape as **w**, whose value (together with **w**) is retained from iteration to iteration. Before the first iteration, $w_{aij} = 0$, $W_{aij} = -w_{min}(w_{max} - w_{min})/w_{max}$.

The algorithm iteration with the image $\mathbf{g} = g_{ij}$ ($\mathbf{g} \in \mathbf{G}$) processed is described by the following pseudo-code:

```
c ← the tensor of convolutions of g with w with the stride s.
E ← ∅
while max(c) > 0 & |E| < N_C
    <a, p, q> ← indices of random element of c such that c_apq = max(c)
    if a ∉ E
        E ← E ∪ {a}
        g^w ← K×K tile of g with upper left corner coordinates (ps, qs)
        if max(c) < 1
            n_b ← the number of pixels in g^w brighter than B.
            if 0 < n_b < K^2
                for all 0 ≤ i, j < K
                    W_aij ← { W_aij + l       if g^w_ij > B
                            { W_aij - ln_b/(K^2 - n_b)   o/w
                    w_aij ← w(W_aij)
                end for
            end if
        else
            p ← 0, n_b ← 0, k ← 0
            r ← set of all values in g^w in decreasing order
            while p < 1
                for all 0 ≤ i, j < K
                    if g^w_ij = r_k
                        p ← p + r_k w_aij
                        W_aij ← W_aij + l
                        w_aij ← w(W_aij)
                        n_b ← n_b + 1
                    end if
                end for
                k ← k + 1
            end while
            for all 0 ≤ i, j < K
                if g^w_ij ≤ r_k
                    W_aij ← W_aij - ln_b/(K^2 - n_b)
                    w_aij ← w(W_aij)
                end if
            end for
        end if
    c_brt ← 0 for all <b, r, t> for which r = p & t = q
end while
```

In this algorithm,

$$w(W) = w_{min} + \frac{(w_{max} - w_{min}) \max(W, 0)}{w_{max} - w_{min} + \max(W, 0)}. \quad (1)$$

Since the algorithm is not very simple, it needs an informal clarification. It can be understood in terms of a network with $N_C$ layers of neurons. The neurons have square receptive fields of the size $K \times K$ corresponding to the convolution matrices overlaid on the input image with the stride $s$. The matrix of synaptic weights of all neurons in the layer $a$ is the same and equals to $w_{aij}$. A step of the algorithm above may be considered as an action of synaptic plasticity mechanism but applied to the whole layer instead of individual neurons. More exactly, the plasticity rule is applied to an auxiliary variable characterizing synaptic strength - $W_{aij}$. The synaptic weight depends on this variable in accordance with (1). $W_{aij}$ corresponds to *synaptic resource* from [17 – 19]. This plasticity mechanism acts as following. The tensor **c** can be interpreted as stimulation values (= membrane potentials) of all neurons. The algorithm finds the most stimulated neurons ("winners") on each layer. It begins with the most stimulated winner and does the same things with all winners. If the winner stimulation is low (< 1), the resources of all its synapses connected to bright pixels are increased. It gives the winner better chance to get higher stimulation from a similar input image next time. If the stimulation is high ($\geq 1$), then the smallest set of the brightest pixels which give the total stimulation $\geq 1$ is determined. All the respective synapses are strengthened. It is important that the total resource of each neuron $\sum_{i,j} W_{aij}$ remains constant. It means that whenever some $W_{aij}$ are increased, all other $W_{aij}$ (for the same $a$) are equally decreased. It introduces competition between synapses. Besides that, the imaginary neurons corresponding to the same input image patch on the different layers are mutually connected by strong inhibitory links prohibiting winners in the different layers to occupy the same place on the input image (WTA - winner-takes-all rule).

Thus, our algorithm resembles the unsupervised learning algorithm described in [10].

All matrices in **w** are significantly different because if some of them were similar then winners on the different layers would correspond to the same location on the image but it is prohibited by the WTA rule.

Matrices in **w** represent typical structural elements in the image set **G**. Since these patterns are frequent, they cause the fastest growth of the respective synaptic weights in the winners making them win more and more frequently.

### III. PORTING THE CONVOLUTION MATRICES INTO THE CONVOLUTIONAL SNN

The overall architecture of convolutional SNN is similar to convolutional ANN – the alternation of convolutional and pooling layers.

The structure of a convolutional layer is totally identical to ANN. It includes several maps consisting of neurons with square receptive fields and identical (for neurons from one map) matrices of synaptic weights. These weight matrices are the matrices **w** multiplied by a constant regulating the mean firing frequency of the convolutional neurons.

The structure of pooling layers depends on the pooling operation and the form of encoding of numeric values by spikes. In [10], the latency coding was selected. In this case, the max pooling is implemented trivially – the pooling neuron simple propagates the first spike from the convolutional neurons ignoring the other spikes. We use the rate-based coding. In this coding scheme, the average pooling has the simplest implementation – by a neuron which propagates all its presynaptic spikes.

To guarantee correct average pooling we slightly modified the neuron model. It is the standard intergrate-and-fire (IF) neuron with current-based delta-synapses but with one distinction. When a neuron fires, its membrane potential $u$ is not reset to the rest potential value ($=0$) – instead the value of the neuron's threshold potential $u_{THR}$ is subtracted from it. This feature guarantees that the count of spikes emitted by a neuron is exactly equal to the count of incoming spikes if synaptic weights just slightly exceed $u_{THR}$. Thus, the change of $u$ during one emulation step is described by the following formulae:

$$u \leftarrow u\left(1 - \frac{1}{\tau}\right) + \sum w_i$$
$$if\ u \geq u_{THR}, u \leftarrow u - u_{THR}, firing \quad (2)$$

Here, $\tau$ is the membrane potential decay time constant, $w_i$ – the synaptic weights (only the weights of the synapses obtaining a spike are summed). For pooling layer neurons, $\tau = 1$. For the convolution layer, $\tau$ should be equal to the image presentation time because neurons there should integrate all spikes from one image. Since we use rate-based coding, one image should be presented for a certain time necessary to express pixel brightness. The SNN CoLaNET [20], used as a classifier in our network, assumes that presentation of one image takes 10 time quants of SNN emulation (further, as in many SNN-related works, we assume that 1 quant = 1 msec), after which 10 msec of silence follow. This 10 msec time gap between images is necessary to prevent influence of previous image due to remaining non-zero membrane potentials. Thus, the pixel brightness is linearly projected to the spike count from the interval [0, 10], and $\tau = 10$.

It should be noted that max pooling for rate coding can also be implemented but this implementation is not exact. It uses the same mechanism as in the case of latency coding and assumes that for several Poissonian processes, the first spike (the spike with the minimum latency) most probably belongs to the process with the maximum frequency. Therefore, max pooling can be implemented by a "winner-takes-all" neuron ensemble where the neuron firing first blocks all other neurons (while the winner itself can continue firing).

### IV. USING THE CONVOLUTIONAL SNN FOR OBJECT CLASSIFICATION IN THE NEOVISION2 DATASET

Neovision2/Tower [21] is a widely used dataset for comparative evaluation of algorithms for object detection and recognition in video streams. It contains frames from short video clips filmed from a fixed camera on top of the Stanford Hoover tower.

Objects in the Neovision2 dataset are marked by rectangles of different sizes and orientations. In order to feed these rectangular images to one fixed SNN, we had to normalize them - to transform to the same shape. We selected square shape 31×31 pixels. The majority of objects in Neovision2 are relatively small – not much bigger than 31×31. Besides that, the classifying SNN CoLaNET [20] is implemented most efficiently when its input node count does not exceed 1024 that limits the size of the images classified. In order to make the classification invariant with respect to brightness/contrast changes and to amplify structural properties of the images, we convert the images to "heat maps" reflecting local pixel brightness variations using the algorithm described below.

Thus, we performed the following manipulations with the original Neovision2 images (Fig. 1):

1. The images were made monochrome.
2. For each marked object, the smallest square containing it with vertical and horizontal sides was found (Fig. 1a). Only the clearly seen objects were selected, not occluded by anything and not laid over other objects. For example, only 4 non-intersecting objects on Fig. 1a are selected.
3. The differential "heat map" is created using the simplest differential algorithm: every pixel of the heat map is calculated as

$$h_{i,j} = \sqrt{(p_{i-1,j} - p_{i,j})^2 + (p_{i+1,j} - p_{i,j})^2 + (p_{i,j-1} - p_{i,j})^2 + (p_{i,j+1} - p_{i,j})^2},$$

where $p_{i,j}$ – the pixels of the original picture. After that, the pixel brightness is scaled to [0, 255] interval (Fig. 1b).
4. The selected square heat map patches are extracted as separate images and shrinked to 31×31. Shrinking was done by taking the brightest pixel from the original pixels projecting to the given pixel of the resulted image. As a result, a set of 31×31 images was obtained (Fig. 1c).

This procedure was applied to the clips from the Neovision2 dataset that gave 155881 images. They were randomly shuffled. In such a way, we created the set **G** from the previous section.

In this study, we tested one-layer convolutional SNN (one convolution layer + one pooling layer). We planned to use 28

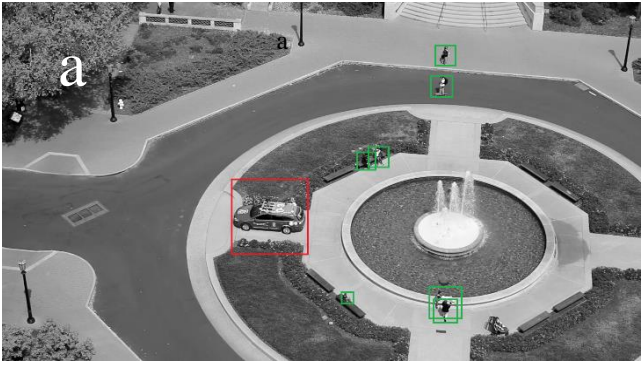
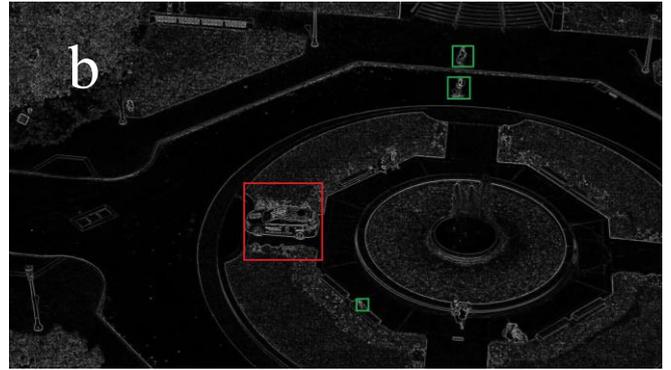
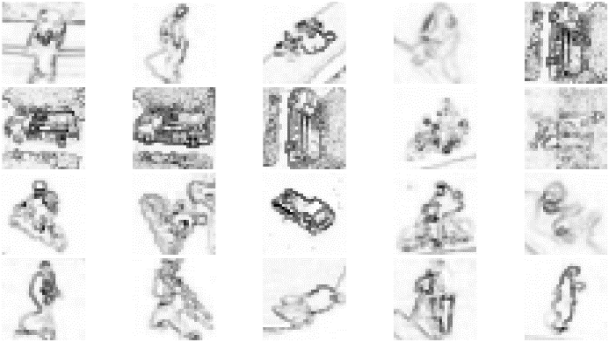

Fig. 1. Creation of image set for obtaining the convolution matrices. a. An orginal image converted to monochrome with square object labels. b. The differential heat map. c. Examples of images used to obtain the convolution matrices (here, brightness is inverted).

maps of 9×9 convolutions with stride 2. It gives convolutions of size 12×12. After 2×2 pooling we obtain 28 6×6 matrices, whose elements correspond to 1008 flattened input nodes of the classifying SNN (as it was said, we intended to avoid exceeding 1024 input node count for performance reasons). Thus, in our algorithm described earlier: $K = 9$, $N_C = 28$, $s = 2$. As the brightness threshold $B$, we took the mean of pixel brightness in **G** (26). Choice of the learning rate $l$ is a matter of trade-off. Lower learning rate leads to slower but more accurate learning. However, learning should be sufficiently fast to guarantee that the correct weight values can be reached during presentation of learning examples. Therefore, it should depend on the learning set size $N_E$. Learning should be able to produce significantly great weights. In case of our algorithm a weight is great if it can allow a signal from one pixel to cause neuron firing. Considering possible negative contribution of other pixels, its contribution should be several times greater than the threshold. Since the mean pixel brightness is about 30, we selected $l = 100 / 255 / N_E \sim$ 8e-6. Choice of the proper values of $w_{min}$ and $w_{max}$ is least evident. Since the number of the pixels lighter than $B$ is approximately 3 times less than the darker pixels, positive weights will be also 3 times fewer.

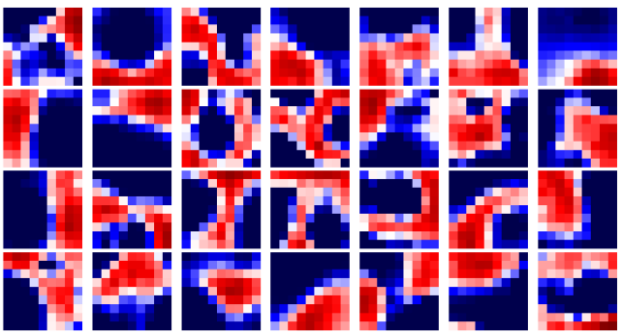

Fig. 2. Color coded values of the convolution matrices.

Respectively, for sake of balance, it is reasonable to make $w_{max}$ 3 times greater than the absolute value of $w_{min}$. On the other hand, as it was said, the contribution of one pixel should be made potentially greater than the threshold. Taking this into account, we selected: $w_{max} = 5 / 255 = 0.0196$, $w_{min} = -5 / 3 / 255 = 0.006536$.

Now, all the necessary data for our algorithm is present. Its result is shown on Fig. 2. Here, the values of 28 square matrices are coded by color – blue corresponds to negative values, red – to positive. We observe formation of very interesting structures – spots, circles, boundaries etc.

As it was said, the procedure of porting the obtained convolutional matrices into SNN is quite straightforward. The sets of synaptic weights of all neurons corresponding to the same feature map are identical and proportional to elements of their convolutional matrix. The proportionality coefficient is determined from the following consideration. CoLaNET learns efficiently if the mean input spike frequency is about 1 spike per input per image. Since the image presentation period is 20 msec, the proportionality coefficient should get the value giving the mean pooling layer firing frequency close to 50 Hz. This value can be easily found by a simple calibration procedure using the fact that the dependence of the output frequency on the input stimulation is monotonous and almost linear in the network working regime.

The resulting whole network structure is depicted on Fig. 3.

Before considering the results, demonstrated by this network on the Neovision2 dataset, we should say some words about CoLaNET, the network performing classification itself. Since it is described in detail in the publicly available source [20], and it is not the main topic of the present paper, we just outline its general idea.

The CoLaNET architecture is presented on Fig. 4. It consists of columns and layers (the name CoLaNET means "columnar layered network"). One column corresponds to one target classs – therefore, in our case CoLaNET includes 5 columns: "person", "cyclist", "car", "bus", "truck". Each column incudes several triplets of neurons ("microcolumns") recognizing significantly different instances of a target class (Fig. 4b). Each layer includes neurons with specific functionality: L - learning neurons; WTA neurons which mutually block each other preventing recognition of similar instances by different microcolumns; REWGATE neurons passing reward signals to an active L neuron; BIASGATE neurons, the sources of additional stimulation in case when L neurons do not fire by themselves; OUT neurons which indicate the target class recognized.

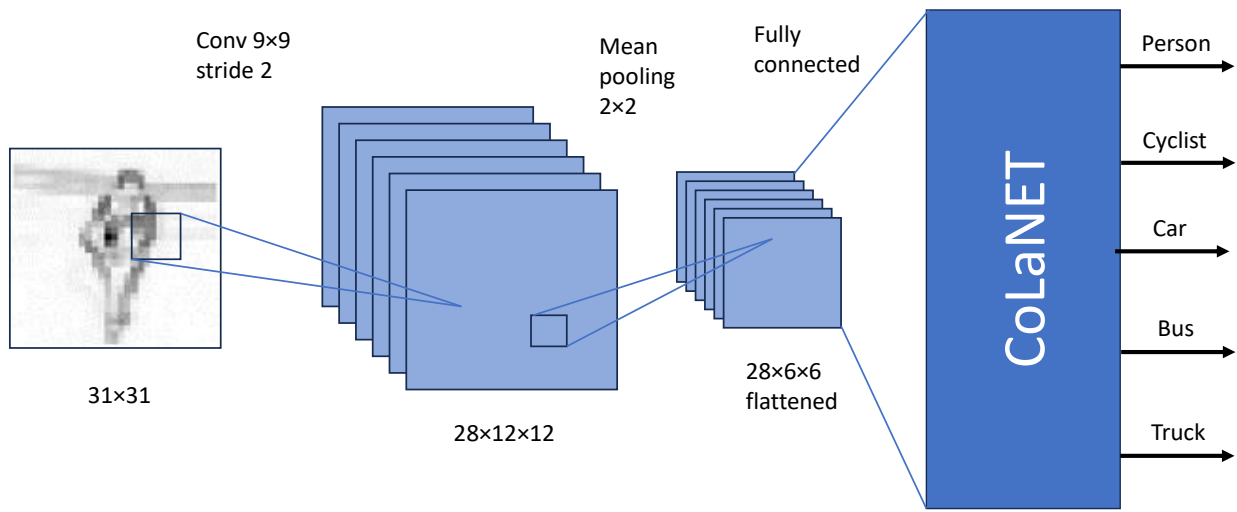

Fig. 3. Color coded values of the convolution matrices.

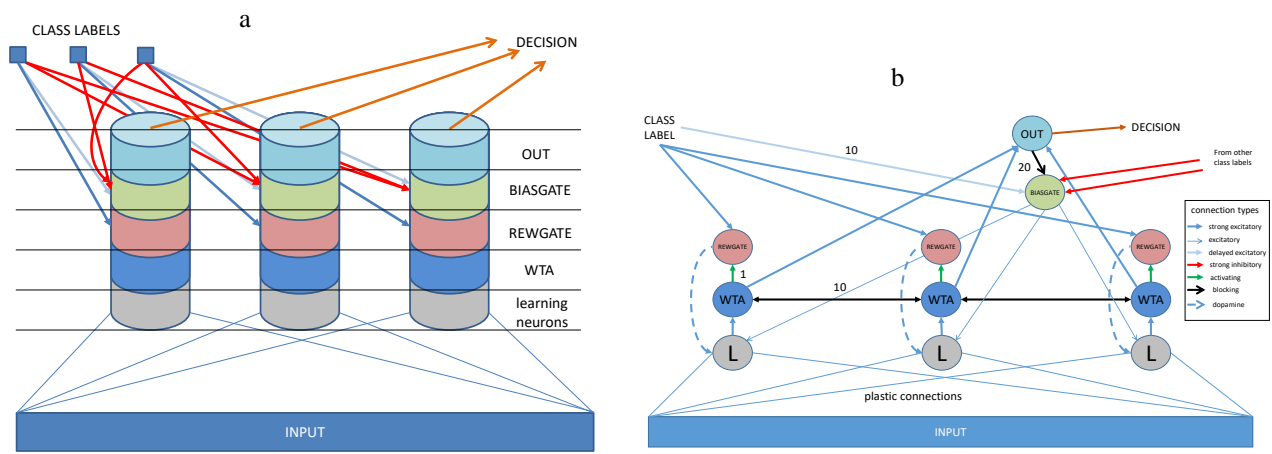

Fig. 4. CoLaNET architecture. a. General view. b. Structure of one column.

Learning is based on a combination of anti-Hebbian and dopamine plasticity. Anti-Hebbian mechanism "punishes" firing L neurons. However, if an L neuron fired correctly – during a presentation of its target class, it will obtain "reward" – the spike coming to its special "dopamine" synapse which will over-compensate the previous suppression resulting in potentiation of those synapses which contributed to firing.

The other components of the CoLaNET structure serve to promote learning in the beginning, when synapses of L neurons are insufficiently strong for firing, and to keep the microcolumn diversity necessary for the column to cover all possible particular cases of the target class.

The details of CoLaNET functioning, setting its hyperparameters and other questions are described in [20] and will not be considered here. Instead, let us discuss accuracy demonstrated by our convolutional SNN (CSNN) on the Neovision2 dataset in comparison with traditional convolutional neural networks (CNN).

For this comparative study, we used Keras package. We selected two following CNN configurations.

Since our task has some similarity with the popular MNIST dataset (many small b/w images) we selected a CNN configuration close to the respective example from Keras web site (https://keras.io/examples/vision/mnist_convnet/) – CNN1. This network shows 99% accuracy on MNIST. It consists of the following layers:

1. convolutional 32×4×4, stride 1, activation - ReLu
2. max pooling 2×2
3. convolutional 64×3×3, stride 1, activation - ReLu
4. max pooling 2×2
5. flatten, dropout 0.5
6. fully connected, activation – softmax, 5 target classes

We selected the second configuration CNN2 most similar to ours:

1. convolutional 28×9×9, stride 2, activation - ReLu
2. max pooling 2×2
3. flatten, dropout 0.5
4. fully connected, activation – softmax, 5 target classes

For the both CNNs, we selected the batch size equal to 128 (for CSNN this parameter has no sense). Since SNNs will be most probably used for online learning without enough memory for storing the whole training set, we used one epoch learning for all the networks. The other parameters for the CNNs were: loss function – categorical crossentropy, optimizer – adam. The CoLaNET parameters: microcolumn count per column 22, learning rate 0.0035, minimum weight -0.0628, maximum weight 0.152. 5-fold cross-validation was used to evaluate accuracy for all networks.

The results are presented in Table 1.

TABLE I. Classification Accuracy for Neovision2 Dataset

| Network | Accuracy, % | |
| --- | --- | --- |
| | *mean* | *standard deviation* |
| CNN1 | 94.35 | 0.32 |
| CNN2 | 92.34 | 0.42 |
| CSNN | 91.58 | 1.1 |

## V. Discussion of the Results

From Table 1 we conclude that CSNN is slightly less accurate than usual convolutional networks. It is rather common situation – in many tests SNNs appear to be less accurate than ANNs. It is explained among other factors by more rough encoding of input information by spikes than by numbers. In our case, instead of 256 pixel brightness degrees, only 11 are used (0-10 spikes per image presentation). However, SNNs have strong advantage in terms of energy consumption. It is demonstrated in many tests [22] that SNNs consume several orders of magnitude less energy than ANNs in similar tasks. This factor can potentially overweigh the moderate loss in accuracy. And we see that CNN1, which is noticeably more accurate than CSNN, is also several times bigger than it in terms of neuron count – 42885 against 9412.

## VI. Conclusion

One of the most promising application areas of modern and future neuroprocessors designed for SNN emulation is processing and analysis of multimedia data. But efficient solution of these problems seems to be unthinkable without utilization of the principle of convolutional networks. Implementation of SNN on neuroprocessors is efficient if only local operations are involved – those which do not require permanent sharing information between numerous neuroprocessor's cores. Training convolutional networks contradicts this idea since it is based on weight sharing - if every convolutional matrix with its own receptive field is implemented by a separate neuron then learning this matrix requires the identical modification of synaptic weights in many neurons simultaneously.

This contradiction motivated us to develop a procedure of obtaining convolutional matrices which would be implemented outside the network – with the subsequent porting these matrices into a convolutional SNN section in the form of weights of non-plastic synapses.

In the present paper we describe this procedure and process of forming convolutional SNN layers using the matrices obtained. We test this whole technology on a practical problem of classification of image fragments obtained from an outdoor camera. We demonstrate that the convolutional SNN created using our approach shows satisfactory accuracy comparable with traditional convolutional ANNs.

Since we implemented our SNN on GPU we cannot measure the CSNN energy consumptions gain in comparison with CNN. But we expect that this gain can be very significant and are going to measure it exactly when we will implement out network on the AltAI neurochip [15]. This work is ongoing now.

In general, we believe that our work is a significant step in widening the neurochip application area.

The supplementary materials for this study are available at https://disk.yandex.ru/d/RijSs6Trc6M_4A.

## Acknowledgment

All computations for this research project were performed on Kaspersky's GPU server equipped with 4 NVIDIA Titan Xp cards. The SNN emulators ArNI-X [23] (owned by M, Kiselev) and Kaspersky Neuromorphic Platform (owned by Kaspersky) were used.

We would like to thank Artem Vorontsov for many interesting ideas related to this study.